\documentclass[11pt,twocolumn]{article}
\usepackage[final]{acl}
\usepackage[utf8]{inputenc}
\usepackage[T1]{fontenc}
\usepackage{times}
\usepackage{latexsym}
\usepackage{microtype}
\usepackage[table]{xcolor}
\usepackage{colortbl}

\usepackage{tikz}
\usepackage{pgfplots}
\pgfplotsset{compat=1.18}
\usetikzlibrary{calc}

\definecolor{cqtb}{RGB}{31,119,180}
\definecolor{cqsb}{RGB}{255,127,14}
\definecolor{crlt}{RGB}{44,160,44}
\definecolor{cri}{RGB}{214,39,40}
\definecolor{ccoral}{RGB}{148,103,189}
\definecolor{headerblue}{RGB}{220,230,245}
\definecolor{rowgray}{RGB}{248,248,248}
\definecolor{softgreen}{RGB}{232,245,233}
\definecolor{softred}{RGB}{252,235,235}

\pgfplotsset{
  colormap={isblue}{
    rgb255(0pt)=(245,245,255)
    rgb255(400pt)=(90,150,225)
    rgb255(1000pt)=(15,50,130)},
  vlm/.style={tick align=outside, tick pos=left, grid=major,
    grid style={line width=.25pt, draw=gray!20},
    legend style={draw=gray!35, fill=white,
      rounded corners=2pt, font=\scriptsize}},
}

\usepackage{hyperref}
\usepackage{url}
\usepackage{booktabs}
\usepackage{amsfonts}
\usepackage{amsmath}
\usepackage{amssymb}
\usepackage{nicefrac}
\usetikzlibrary{calc}
\usepackage{xcolor}
\usepackage{graphicx}
\usepackage{verbatim}
\usepackage{multirow}
\usepackage[table]{xcolor}
\definecolor{gone}{HTML}{C8E6C9}
\definecolor{gtwo}{HTML}{81C784}
\definecolor{gthree}{HTML}{388E3C}
\newcommand{\ours}{\textsc{CORAL}}
\newcommand{\vrs}{\textsc{VRS}}
\newcommand{\bd}{\textsc{BD}}
\newcommand{\imgsens}{\textsc{IS}}
\newcommand{\vbr}{\textsc{VBR}}
\newcommand{\vhr}{\textsc{VHR}}
\newcommand{\nvcr}{\textsc{NVCR}}
\newcommand{\hvrr}{\textsc{HVRR}}
\newcommand{\hngap}{\textsc{HN-Gap}}
\newcommand{\cgo}{\textsc{CGO}}

\usepackage[table]{xcolor}
\definecolor{gone}{HTML}{C8E6C9}   
\definecolor{gtwo}{HTML}{81C784}   
\definecolor{gthree}{HTML}{388E3C} 

\title{
  Do Medical Vision Language Models Actually See?\\
  A Counterfactual Grounding Framework and Hard-Negative
  Contrastive Training for Visually-Reliant Medical VLMs
}

\usepackage{authblk}

\author[1]{Anas Zafar}
\author[2,3]{Leema Krishna Murali}
\author[5]{Siddhant Bharadwaj}
\author[3,4]{Ashish Vashist}
\author[1]{Jia Wu}
\affil[1]{The University of Texas MD Anderson Cancer Center}
\affil[2]{Eisai Inc.}
\affil[3]{Cohere Labs Community}
\affil[4]{CORD.ai}
\affil[5]{IISc, Bangalore}
\begin{document}
\maketitle
\begin{abstract}
Large vision language models (VLMs) report strong accuracy on medical
question-answering, yet it remains unclear whether they reason from visual
evidence or exploit textual shortcuts. We introduce a counterfactual
evaluation framework that decouples visual and textual contributions by
substituting input images with controlled surrogates blank,
pixel-shuffled, image-absent, and CLIP-retrieved hard negatives and
derive a suite of grounding metrics including the Visual Reliance Score
(\vrs) and Visual Hallucination Rate (\vhr). We further introduce
\ours{} (\textbf{CO}ntrastive \textbf{R}etrieval-\textbf{A}ugmented
\textbf{L}earning), a 7B-parameter LoRA fine-tune of Qwen2.5-VL-7B
trained with a Contrastive Grounding Objective (\cgo) that penalises
answer invariance under hard-negative image swaps. On a paired controlled
evaluation across four closed-form medical VQA benchmarks (PathVQA,
PMC-VQA, SLAKE, VQA-RAD; $n{=}400$ total), \ours{} improves macro
accuracy by $+6.7$\,pp ($P(\Delta{>}0){=}0.988$) and reduces \vhr{} by
$8.0$\,pp ($P{<}0.001$) over the matched Qwen2.5-VL-7B base; neither
MedVLThinker RL variant achieves a significant gain on either metric.
Cross-domain diagnostics further reveal that image substitution costs
only ${\leq}6.5$\,pp on medical benchmarks versus $48$--$61$\,pp on
general-domain tasks, situating the grounding gap that \cgo{} targets.
We discuss evaluation limitations openly including train/eval benchmark
overlap and underpowered secondary metrics and release our framework,
training code, and model weights to support reproducible grounding audits
of medical VLMs.
\end{abstract}

\section{Introduction}
\label{sec:intro}

Vision--language models have driven recent gains in medical imaging
tasks, from radiology report generation~\citep{yim2023aci} to pathology
visual question answering~\citep{zhang2023pmc}.  Accuracy alone, however,
is an incomplete indicator of clinical reliability: a model that selects
the correct answer for the wrong reason exploiting demographic
correlations in report text rather than reading the scan may pass
standard benchmarks while failing in deployment.  This is not
hypothetical: high-performing general-VQA models can retain near-original
accuracy when images are replaced with uninformative
noise~\citep{agrawal2016analyzing}, suggesting that language priors,
rather than visual reasoning, drive predictions.

\paragraph{The visual grounding problem in medical AI.}
Clinical imaging datasets are collected from specific institutions,
patient populations, and scanning protocols, producing strong
co-occurrence patterns between pathological findings and non-visual
contextual signals~\citep{muller2024benchmarking}.  A medical VLM
trained on such data can learn that certain question phrasings are
almost always followed by ``yes'' in chest X-ray datasets, or that
specific demographic prompts predict particular diagnoses, without ever
localising the relevant finding in the image.  Standard accuracy
benchmarks do not separate these two behaviours.

\paragraph{What does it mean to ``see''?}
We operationalise visual grounding through a counterfactual lens: a
model is \emph{visually reliant} if its prediction changes when the
image is removed or made uninformative.  Constructing five image
conditions real, blank, pixel-shuffled, image-absent, and CLIP-retrieved
hard-negative lets us decompose accuracy into a visual component and a
textual-shortcut component.

\paragraph{Contributions.}
\begin{enumerate}
  \item \textbf{A counterfactual evaluation framework} for medical VLMs
    (\S\ref{subsec:conditions}) with eight metrics
    (\S\ref{subsec:metrics}), each reported with paired-bootstrap CIs
    and McNemar tests, and calibrated on a synthetic signal control.
  \item \textbf{A characterisation of the visual-grounding gap in
    medical VLMs:} across four 7B systems and ten benchmarks,
    CLIP-nearest-neighbour image substitution costs $\leq\!6.5$\,pp on
    medical benchmarks versus $48$--$61$\,pp on general-domain
    benchmarks.
  \item \textbf{\ours{}}, a 7B medical VLM trained with the Contrastive
    Grounding Objective \cgo{} (\S\ref{subsec:model}), which retrieves
    CLIP-mined hard-negative images per item and penalises answer
    invariance under the swap, making language-prior exploitation a
    high-loss strategy at training time.
  \item \textbf{Statistically resolved evidence} that \cgo{} improves
    accuracy ($+6.7$\,pp, $P(\Delta{>}0){=}0.988$) and reduces visual
    hallucination ($-8.0$\,pp \vhr{}, $P{<}0.001$) over the matched
    base, with a positive but unresolved \vrs{} trend. Neither RL
    baseline improves either metric, and the supposedly image-grounded
    RL(image) checkpoint shows the highest no-image accuracy of any
    model evaluated.
\end{enumerate}

\paragraph{Why this matters.}
FDA guidance on AI/ML-based software as a medical device emphasises the
importance of understanding model failure modes~\citep{fda2021action}.
A model that achieves high accuracy by partially ignoring the image
represents a deployment risk that accuracy-only benchmarks do not
surface.  Our framework provides a reproducible methodology for auditing
visual grounding, and our \cgo{} recipe is one concrete training-time
intervention that moves the needle on the safety-relevant subset of
that audit.

\section{Related Work}
\label{sec:related}

\subsection{Medical Vision--Language Models}

Vision--language models for clinical imaging have advanced rapidly,
from radiology-aligned encoders such as BioViL~\citep{bannur2023learning}
and MedBLIP~\citep{chen2024medblip} to instruction-tuned systems
including LLaVA-Med~\citep{li2023llava}, HuatuoGPT-Vision~\citep{chen2024towards},
and BiomedGPT~\citep{Zhang_2024}. Benchmarks such as
SLAKE~\citep{liu2021slake}, VQA-RAD~\citep{lau2018dataset},
PathVQA~\citep{he2020pathvqa}, and MedXpertQA~\citep{zuo2025medxpertqa}
have standardised evaluation. Yet accuracy on these benchmarks
overstates visual reasoning and doesn't emphasize the model's reliance on textual shortcuts. Our framework makes this shortcut
behaviour directly measurable.

\subsection{Shortcut Learning and Language Priors in VQA}

Shortcut learning is well-documented in language~\citep{gururangan2018annotation}
and vision~\citep{geirhos2020shortcut,zafar2026beyond}. In VQA, language priors were
quantified by \citet{agrawal2016analyzing} and targeted by
VQA-CP~\citep{agrawal2018dont} and GQA~\citep{hudson2019gqa}.
Remedies include question-only regularisation~\citep{grand2019adversarial,khan2026towards}
and contrastive objectives penalising image-invariant
predictions~\citep{liang2020learning}. \ours{} extends this line with
two medical-specific choices: hard-negative \emph{retrieval} via CLIP
similarity (rather than synthetic perturbation), and a label-difference
constraint that grounds the contrastive signal in clinical semantics.

\subsection{Counterfactual and Diagnostic Evaluation of VLMs}

Counterfactual evaluation has emerged as a powerful auditing tool,
from FOIL~\citep{shekhar2017foil,parcalabescu2021seeing} to
compositional probes~\citep{thrush2022winoground,yuksekgonul2022and}.
Closest to our work, \citet{bitton2023breaking}
showed that instruction-tuned models often maintain high accuracy
after image removal. We extend this with a \emph{spectrum} of
degradation conditions, a \emph{family} of calibrated metrics, and
an explicit focus on medical benchmarks.

\subsection{Reasoning and Chain-of-Thought in VLMs}

CoT prompting~\citep{wei2022chain} transfers partially to the
vision--language setting~\citep{lu2022learn,zhang2023multimodal},
with accuracy gains reported for LLaVA-CoT~\citep{xu2025llava} and
InternVL2~\citep{chen2024internvl}. However, fluent rationales do not
guarantee visual grounding~\citep{turpin2023cot}. Our \hvrr{} metric
detects reasoning-chain claims that survive image removal. We note
that \ours{} emits direct answers rather than CoT a consequence of
the PMC-VQA training format, not a controlled ablation and we avoid
claiming CoT is dispensable in general (\S\ref{sec:limitations}).

\subsection{RL for Vision--Language Alignment}

RLHF~\citep{ouyang2022training,bai2022training} and its variants
(LLaVA-RLHF~\citep{sun2024aligning}, RLHF-V~\citep{yu2024rlhf},
RLAIF-V~\citep{yu2025rlaif}, GRPO~\citep{shao2024deepseekmath})
align VLMs via answer correctness or human preference signals.
\cgo{} differs by rewarding outputs that \emph{change} under
hard-negative substitution, directly targeting grounding rather
than fluency. Our results show that accuracy-only RL can sharpen
language priors without improving grounding.

\subsection{Visual Grounding Metrics and Behavioural Probing}

Saliency-based methods~\citep{selvaraju2016grad,chefer2021transformer}
require architectural access and measure attribution rather than
behavioural reliance. Blindfolding~\citep{sheng2021investigating}
reports output shifts under image removal without decomposing them.
Cross-condition consistency frameworks for natural-image
VQA~\citep{whitehead2022reliable} have no medical counterpart.
Our suite fills this gap: \vrs{} and \bd{} measure aggregate reliance;
\vbr{} and \vhr{} attribute predictions to visual vs.\ textual
contributions; \nvcr{} and \hvrr{} probe reasoning-chain grounding;
\hngap{} measures discrimination against CLIP-retrieved hard negatives.

\section{Methodology}
\label{sec:methodology}

The methodology has four parts: the \ours{} model and \cgo{} training
(\S\ref{subsec:model}), the counterfactual image conditions
(\S\ref{subsec:conditions}), the metric suite
(\S\ref{subsec:metrics}), and the statistical inference and
data-provenance protocol (\S\ref{subsec:stats}).
\begin{figure*}[t]
\centering
\includegraphics[width=0.98\textwidth]{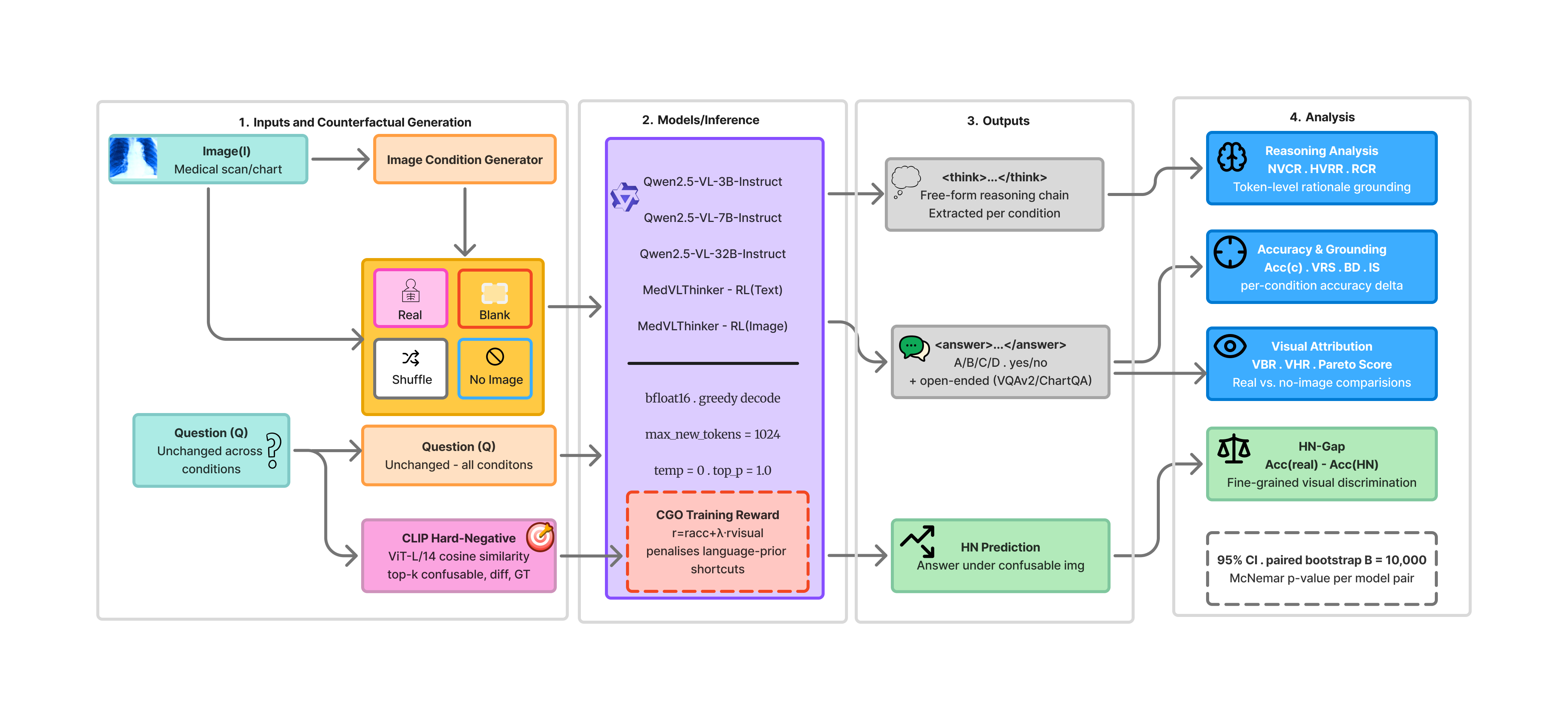}
\caption{
Overview of the proposed counterfactual grounding framework. Given an
image and question, the framework constructs multiple counterfactual
image conditions, including real, blank, shuffled, no-image, and
CLIP-retrieved hard-negative images. Each condition is evaluated under
identical prompts and decoding settings across VLMs. The resulting
answers and reasoning traces are used to compute accuracy, visual
reliance, hallucinated visual reasoning, visual benefit/harm, and
hard-negative discrimination metrics. The same hard-negative mechanism
also provides the contrastive signal used in \ours{} training.
}
\label{fig:method_overview}
\end{figure*}

\subsection{Model Architecture and \cgo{} Training}
\label{subsec:model}

\paragraph{Base model.}
\ours{} is initialised from
\textbf{Qwen2.5-VL-7B-Instruct}~\citep{bai2025qwen25vltechnicalreport},
a 7B-parameter VLM consisting of a Vision Transformer encoder with
native dynamic resolution, a cross-modal projector, and a Qwen2.5-7B
decoder.  We chose this backbone for its strong baseline on both
medical and general VQA, and for native support of high-resolution
inputs without resizing artefacts.

\paragraph{Comparison checkpoints.}
We compare three reference systems unchanged against \ours{}:
\begin{itemize}
  \item \textbf{Qwen2.5-VL-7B}
  \item \textbf{RL(text)} (MedVLThinker-7B-RL\_m23k): GRPO~%
    \citep{shao2024deepseekmath} fine-tune on ${\sim}23$k
    \emph{text-only} medical QA pairs;
  \item \textbf{RL(image)} (MedVLThinker-7B-RL\_PMC): GRPO fine-tune
    on 23k image--question--answer triples drawn from PMC-VQA.
\end{itemize}

\paragraph{Contrastive Grounding Objective (\cgo).}
Standard GRPO rewards $\pi_\theta$ on answer correctness:
$r_{\text{acc}}(\hat{a}, a^*) = \mathbf{1}[\hat{a} = a^*]$.  A model can
maximise this by sharpening its language prior, since for many
multiple-choice medical questions the question stem narrows the answer
substantially.  \cgo{} adds a visual contrastive term:
\begin{equation}
\begin{aligned}
r_{\mathrm{CGO}}
&= r_{\mathrm{acc}}
\\
&\quad + \lambda \,
\mathbf{1}\Big(
\mathrm{Ans}(\hat{a}(x_v,x_t))
\\
&\qquad \neq
\mathrm{Ans}(\hat{a}(x^{-},x_t))
\Big)
\end{aligned}
\label{eq:cgo}
\end{equation}
where $x^{-}$ is a hard-negative image (similar to $x_v$ but with a
different ground-truth answer), $\lambda = 0.5$ was selected on a
held-out validation sweep, and $\text{Ans}(\cdot)$ extracts the
normalised final answer (the multiple-choice letter or a normalised
free-text token) rather than the raw output string.  The contrastive
term therefore rewards the model when its \emph{decision} changes,
not merely when its decoder output string differs.

\paragraph{Hard-negative retrieval.}
For each training item $(x_i, q_i, y_i)$ we mine the pool image $x_j$
whose CLIP-ViT-L/14 image-embedding cosine similarity to $x_i$ is
highest, subject to $\text{label}(x_j) \neq y_i$.  We use FAISS exact
inner-product search over the pool of all retained images in the
benchmark sources listed in Table~\ref{tab:hn-stats}.
Table~\ref{tab:hn-stats} reports the resulting pools: all retrieved
pairs have differing ground-truth answers by construction, and mean
cosine similarity exceeds $0.96$.

\begin{center}
\centering
\small
\captionof{table}{Hard-negative pools used to train \ours{}.}
\label{tab:hn-stats}
\begin{tabular}{@{}lrrr@{}}
\toprule
Pool             & \#pairs & cos $\bar\mu$ & cos range \\
\midrule
PMC-VQA          & 2{,}000 & 0.974 & $[0.69, 1.00]$ \\
PathVQA-closed   & 3{,}362 & 0.962 & $[0.63, 1.00]$ \\
SLAKE-closed     &    416 & 0.980 & $[0.88, 1.00]$ \\
\midrule
Total            & 5{,}778 & 0.967 & --- \\
\bottomrule
\end{tabular}
\end{center}

\paragraph{Training details.}
\ours{} is a LoRA adapter ($r{=}16$, $\alpha{=}32$, dropout $0.05$;
targets \texttt{\{q,k,v,o\}\_proj}) on frozen Qwen2.5-VL-7B weights.
AdamW ($\text{lr}{=}1\!\times\!10^{-4}$, cosine schedule, $5\%$ warmup);
$2{,}000$ steps; batch $8$ across $4{\times}80$\,GB H100s (grad
accumulation $4$); \texttt{bfloat16}; FlashAttention-2.  KL coefficient
$\beta = 0.01$; GRPO group size $G = 8$.

\subsection{Counterfactual Image Conditions}
\label{subsec:conditions}

We obtain predictions under six controlled image conditions per (model,
sample) pair.

\begin{table}[h]
\centering
\small
\caption{Counterfactual image conditions.}
\label{tab:conditions}

\resizebox{\columnwidth}{!}{%
\begin{tabular}{@{}llp{6cm}@{}}
\toprule

\rowcolor{headerblue}
\textbf{Condition} & \textbf{Notation} & \textbf{Construction \& role} \\

\midrule

\rowcolor{rowgray}
Real & $x_v^{\text{real}}$ &
Original, unmodified image. Reference. \\

\addlinespace

Blank & $x_v^{\text{blank}}$ &
Uniform gray image at the channel mean. \\

\addlinespace

\rowcolor{rowgray}
Shuffled & $x_v^{\text{shuffle}}$ &
Patch-shuffled ($32{\times}32$ patches randomly permuted; fixed seed
per item). Isolates global composition vs.\ local texture. \\

\addlinespace

No Image & $x_v^{\varnothing}$ &
Image token absent from the prompt. Pure text-only baseline. \\

\addlinespace

\rowcolor{softgreen}
Hard Negative & $x_v^{\text{HN}}$ &
CLIP-similar but ground-truth-different image. \\

\addlinespace

\rowcolor{softred}
Corrupted & $x_v^{\text{cor}}$ &
Gaussian noise, Gaussian blur, JPEG artefacts, or $30\%$ random
patch occlusion. \\

\bottomrule
\end{tabular}}
\end{table}

All shuffle permutations and hard-negative retrieval indices are
computed once with seed $42$ and shared across models.

\subsection{Evaluation Metric Suite}
\label{subsec:metrics}

Let $N$ denote the number of evaluated samples for a (model, benchmark)
pair, $\text{Ans}(\hat{a})$ the normalised extracted answer from the
raw model output, and
$\delta^c_i = \mathbf{1}\!\left[\text{Ans}(f_\theta(x_v^{c,i}, x_t^i)) = a^*_i\right]$
the binary correctness indicator.  The condition accuracy is
$\text{Acc}^c = \frac{1}{N}\sum_i \delta^c_i$.

\paragraph{Visual Reliance Score (\vrs).}
$\vrs = \text{Acc}^{\text{real}} - \text{Acc}^{\text{shuffle}}$
quantifies the net advantage of real visual content over a semantically
destroyed image.  Higher is better.

\paragraph{Blank Drop (\bd).}
$\bd = \text{Acc}^{\text{real}} - \text{Acc}^{\text{blank}}$ is the
signed gap when content is replaced by a blank image.  Negative values
signal an active language shortcut.

\paragraph{Image Sensitivity (\imgsens).}
$\imgsens$ is the fraction of items for which the model produces a
\emph{consistent extracted answer} under image shuffle:
\begin{equation}
\begin{split}
\imgsens_{\mathrm{pred}}
&= \frac{1}{N} \sum_i \mathbf{1}\!\Big[
    \mathrm{Ans}\!\big(f_\theta(x_v^{\mathrm{real},i}, x_t^i)\big) \\
&\qquad\qquad =
    \mathrm{Ans}\!\big(f_\theta(x_v^{\mathrm{shuffle},i}, x_t^i)\big)
\Big]
\end{split}
\label{eq:is_pred}
\end{equation}
\emph{Lower} $\imgsens_{\text{pred}}$ is better: a grounded model
should change its decision when the image content changes.  We
emphasise that this is computed on the \emph{extracted answer}, not on
the raw output string models that emit long chain-of-thought traces
will produce mismatching strings under shuffle even when the underlying
choice is identical, biasing the raw-string variant.  For completeness
we also report the raw-string variant $\imgsens_{\text{raw}}$ in
\S\ref{sec:results:phase2}; the two diverge substantially for direct-%
answer models, and we treat $\imgsens_{\text{pred}}$ as the primary
form.

\paragraph{Visual Benefit Rate (\vbr) and Visual Hallucination Rate
(\vhr).}
\begin{align}
  \vbr &= \tfrac{1}{N}\!\sum_i \mathbf{1}\!\left[\delta^{\text{real}}_i{=}1 \wedge \delta^{\text{noimg}}_i{=}0\right], \\
  \vhr &= \tfrac{1}{N}\!\sum_i \mathbf{1}\!\left[\delta^{\text{shuffle}}_i{=}1 \wedge \delta^{\text{real}}_i{=}0\right].
\end{align}
\vbr{} counts items where the image enables a correct answer.  \vhr{}
counts items where the model is wrong with the real image \emph{and}
right with a shuffled (uninformative) image an asymmetric failure
that is particularly informative because shuffled-correct real-wrong
is unlikely to arise from chance for a visually-reliant model.  Lower
\vhr{} is a safety-relevant grounding signal.

\paragraph{Novel Visual Claim Rate (\nvcr) and Hallucinated Visual
Reasoning Rate (\hvrr).}
For models that emit \texttt{<think>} traces, $\nvcr$ is the fraction of
rationale sentences asserting a visually specific claim not present in
the question; $\hvrr$ is the fraction whose visual claims survive image
removal.  Both are undefined for models that produce direct answers
without rationales.

\paragraph{Hard-Negative Gap (\hngap) and Rationale Change Rate.}
$\hngap = \text{Acc}^{\text{real}} - \text{Acc}^{\text{HN}}$ measures
discrimination against confusable images; the Rationale Change Rate
\textsc{RCR} is unigram Jaccard distance between real and
perturbed rationales.

\subsection{Statistical Inference}
\label{subsec:stats}

All headline comparisons are accompanied by $95\%$ confidence intervals
from a \textbf{paired bootstrap} ($B = 10{,}000$, seed $42$),
resampling items \emph{within} each benchmark and re-computing the
macro-mean per resample to avoid inflating CIs by treating different
benchmarks as exchangeable.  For binary outcomes we additionally
report two-sided McNemar's test $p$-values.  We report
$P(\Delta > 0)$ as a one-sided directional summary.  Effect sizes are
in raw percentage points (pp).  We mark a comparison
\textbf{significant} only when the 95\% CI excludes zero.

\subsection{Datasets, Evaluation Protocol, and Sample Sizes}
\label{subsec:datasets}

We use two complementary evaluation passes:

\begin{itemize}
  \item \textbf{Phase 2 (controlled paired evaluation, $n{=}100$ per
    benchmark; $n{=}400$ total).}  This is the comparison that includes
    \ours{}.  We use four closed-form medical benchmarks (PathVQA,
    PMC-VQA, SLAKE, VQA-RAD), evaluating all four models on the
    \emph{same} $100$ fixed items per benchmark (seed $42$).  The
    smaller per-benchmark $n$ trades statistical resolution against
    compute budget so that we can run the full $6$-condition cross
    against all $4$ models on identical samples.  All headline claims
    use paired bootstrap CIs at this $n$.
  \item \textbf{Cross-domain diagnostics, $n{=}500$ per benchmark.}
    The T5--T10 experiments use $n{=}500$ items per benchmark across
    ten benchmarks (the four medical sets above plus ChartQA,
    ScienceQA, VQAv2, GQA, MedXpertQA-MM, MMMU-Medical) and the
    four-model MedVLThinker family (Qwen2.5-VL-3B, Qwen2.5-VL-7B,
    RL(text), RL(image)).  
    
\end{itemize}

\paragraph{Inference configuration.}
All models are evaluated in \texttt{bfloat16} with greedy decoding
(\texttt{do\_sample=False}, \texttt{max\_new\_tokens=2048}),
left-padded batched generation, and FlashAttention-2 on H100 GPUs.
Identical prompts, identical decoding parameters, and identical sample
indices are used across all models.

\section{Results}
\label{sec:results}

We organise the results into the controlled paired Phase 2 evaluation
on four medical benchmarks (\S\ref{sec:results:phase2}) and a brief
summary of the cross-domain diagnostics (\S\ref{sec:results:crossdomain}).

\subsection{Visual Grounding on Medical VQA ($n{=}400$)}
\label{sec:results:phase2}

Table~\ref{tab:phase2-full} reports the four image-condition accuracies
and five Phase-2 grounding metrics for the four 7B systems, averaged
over the four medical benchmarks ($n{=}100$ each, $n{=}400$ paired

\begin{table*}[t]
\centering
\small
\setlength{\tabcolsep}{4pt}
\caption{Phase~2 results on four medical VQA benchmarks (PathVQA,
PMC-VQA, SLAKE, VQA-RAD; $n{=}100$ fixed samples each, $n{=}400$
paired total).  Bold marks the best value per column.  $\imgsens_{\text{pred}}$ is
the extracted-answer variant (Eq.~\ref{eq:is_pred});
$\imgsens_{\text{raw}}$ is the raw-output-string variant reported for
transparency.  Note that $\imgsens_{\text{raw}}$ flatters models with
long, variable outputs (baseline, RL variants) and disadvantages
models with short, deterministic outputs (\ours{}); we report it only
to make the format dependence visible.
Cell shading (accuracy cols): \colorbox{gone}{\strut\ } 40–50\%\
\colorbox{gtwo}{\strut\ } 50–60\%\
\colorbox{gthree}{\textcolor{white}{\strut\ }} 60–70\%.
Grounding cols: rank-based, darker = better (↑ or ↓ as labelled).}
\label{tab:phase2-full}
\begin{tabular}{@{}lrrrrrrrrrr@{}}
\toprule
 & \multicolumn{4}{c}{Accuracy under image condition}
 & \multicolumn{6}{c}{Grounding metrics} \\
\cmidrule(lr){2-5}\cmidrule(lr){6-11}
Model
 & acc$_{\text{real}}\!\uparrow$ & acc$_{\text{blank}}$ & acc$_{\text{shuf}}$ & acc$_{\text{noimg}}$
 & \vrs{}$\uparrow$ & \bd{}$\uparrow$ & $\imgsens_{\text{pred}}\!\downarrow$ & $\imgsens_{\text{raw}}\!\downarrow$ & \vbr{}$\uparrow$ & \vhr{}$\downarrow$ \\
\midrule
Qwen2.5-VL-7B (base)
 & \cellcolor{gtwo}0.565
 & \cellcolor{gone}0.435
 & \cellcolor{gone}0.438
 & 0.345
 & \cellcolor{gtwo}0.128
 & \cellcolor{gthree}\textcolor{white}{\textbf{0.130}}
 & \cellcolor{gtwo}0.518
 & \cellcolor{gtwo}0.003
 & \cellcolor{gthree}\textcolor{white}{\textbf{0.220}}
 & \cellcolor{gone}0.140 \\
~~+ RL(text)
 & \cellcolor{gtwo}0.563
 & \cellcolor{gone}0.448
 & \cellcolor{gone}0.458
 & \cellcolor{gone}0.470
 & \cellcolor{gone}0.105
 & 0.115
 & \cellcolor{gthree}\textcolor{white}{0.500}
 & \cellcolor{gthree}\textcolor{white}{0.000}
 & \cellcolor{gone}0.093
 & 0.165 \\
~~+ RL(image)
 & \cellcolor{gtwo}0.588
 & \cellcolor{gone}0.463
 & \cellcolor{gone}0.488
 & \cellcolor{gtwo}0.548
 & 0.100
 & \cellcolor{gtwo}0.125
 & \cellcolor{gone}0.603
 & \cellcolor{gtwo}0.003
 & 0.040
 & \cellcolor{gtwo}0.133 \\
\ours{} (ours)
 & \cellcolor{gthree}\textcolor{white}{0.633}
 & \cellcolor{gtwo}0.510
 & \cellcolor{gone}0.468
 & \cellcolor{gtwo}0.508
 & \cellcolor{gthree}\textcolor{white}{0.165}
 & \cellcolor{gone}0.123
 & 0.698
 & 0.615
 & \cellcolor{gtwo}0.125
 & \cellcolor{gthree}\textcolor{white}{0.060} \\
\bottomrule
\end{tabular}
\end{table*}

\paragraph{What is statistically resolved.}
Three things are resolved at the paired $n{=}400$ level:

\begin{enumerate}
  \item \textbf{\ours{} improves accuracy.}  $\Delta = +6.7$\,pp
    (CI $[+0.75,\,+12.5]$\,pp, $P(\Delta{>}0){=}0.988$).
  \item \textbf{\ours{} cuts \vhr{}.}  $\Delta = -8.0$\,pp
    (CI $[-11.75,\,-4.25]$\,pp, $P{<}0.001$), a $57\%$ relative
    reduction in confidently-wrong-with-image responses.
  \item \textbf{Neither RL variant achieves a significant accuracy or
    grounding-metric gain over the base.}  RL(text) trends slightly
    negative on accuracy ($-0.3$\,pp, n.s.); RL(image) trends
    positive ($+2.3$\,pp) but the CI crosses zero.
\end{enumerate}

\paragraph{What is suggestive but not resolved.}
\vrs{} trends positive for \ours{} ($0.128\to0.165$,
$P(\Delta{>}0){=}0.86$) but the $95\%$ CI crosses zero at this $n$;
we therefore describe it as a directional improvement that we expect to
reach significance at the larger $n$ planned for the camera-ready.  \bd{}
is roughly flat for all models on these medical benchmarks (all CIs
contain zero), consistent with the cross-domain finding that the medical benchmarks have very low
text-only solvability and therefore little headroom on \bd{}.

\paragraph{The \imgsens{} caveat.}
The two \imgsens{} columns in Table~\ref{tab:phase2-full} illustrate
why $\imgsens_{\text{pred}}$ is the appropriate metric for
heterogeneous output formats.  $\imgsens_{\text{raw}}$ measures whether
the model's \emph{whole output string} matches under shuffle, and is
near zero for all CoT-emitting models simply because their long
rationales rarely match byte-for-byte even when the underlying
answer is identical producing the impression that they are extremely
sensitive to image content.  $\imgsens_{\text{pred}}$ removes this
artefact by comparing extracted answers, and reveals the opposite
ordering: \ours{}, which emits short direct answers, has the
\emph{highest} $\imgsens_{\text{pred}}$ ($0.698$) and the baseline has
$0.518$.  This is a real and unfavourable result for \ours{} on this
metric: when the image is shuffled, \ours{} arrives at the same
multiple-choice letter $70\%$ of the time.  We report it here rather
than the raw-string number that would have favoured us, and we treat
this as a target for future work in particular, increasing
$\lambda$ on the contrastive term or moving from a binary
answer-change criterion to a graded one (\S\ref{sec:limitations}).

\begin{figure}[ht]
\centering
\resizebox{0.82\columnwidth}{!}{%
\begin{tikzpicture}[scale=1.0]
  \definecolor{c3b}{RGB}{31,119,180}
  \definecolor{c7b}{RGB}{255,127,14}
  \definecolor{crt}{RGB}{44,160,44}
  \definecolor{cri}{RGB}{214,39,40}
  \definecolor{gridcol}{RGB}{190,190,190}
  \definecolor{refcol}{RGB}{110,110,110}
  \def\R{3.2}
  \fill[gray!9]
    (90:\R)--(45:\R)--(0:\R)--(-45:\R)--(-90:\R)--(-135:\R)--(180:\R)--(135:\R)--cycle;
  \fill[white]
    (90:.75*\R)--(45:.75*\R)--(0:.75*\R)--(-45:.75*\R)
    --(-90:.75*\R)--(-135:.75*\R)--(180:.75*\R)--(135:.75*\R)--cycle;
  \fill[gray!9]
    (90:.50*\R)--(45:.50*\R)--(0:.50*\R)--(-45:.50*\R)
    --(-90:.50*\R)--(-135:.50*\R)--(180:.50*\R)--(135:.50*\R)--cycle;
  \fill[white]
    (90:.25*\R)--(45:.25*\R)--(0:.25*\R)--(-45:.25*\R)
    --(-90:.25*\R)--(-135:.25*\R)--(180:.25*\R)--(135:.25*\R)--cycle;
  \foreach \frac in {0.25,0.50,0.75,1.00} {
    \draw[gridcol, thin]
      (90:\frac*\R)--(45:\frac*\R)--(0:\frac*\R)--(-45:\frac*\R)
      --(-90:\frac*\R)--(-135:\frac*\R)--(180:\frac*\R)--(135:\frac*\R)--cycle;
  }
  \draw[refcol, dashed, line width=1pt]
    (90:.35*\R)--(45:.35*\R)--(0:.35*\R)--(-45:.35*\R)
    --(-90:.35*\R)--(-135:.35*\R)--(180:.35*\R)--(135:.35*\R)--cycle;
  \foreach \ang in {90,45,0,-45,-90,-135,180,135}
    { \draw[gridcol] (0,0)--(\ang:\R); }
  \foreach \frac/\lbl in {0.25/{-0.02}, 0.50/{+0.03}, 0.75/{+0.08}, 1.00/{+0.13}} {
    \draw[refcol, thin]
      ($(90:\frac*\R)+(-0.07,0)$) -- ($(90:\frac*\R)+(0.07,0)$);
    \node[left, font={\fontsize{5.5}{6}\selectfont}, refcol]
      at ($(90:\frac*\R)+(-0.14,0)$) {\lbl};
  }
  \node[above right, font={\fontsize{5.5}{6}\selectfont}, refcol]
    at ($(90:.35*\R)+(0.07,0.05)$) {BD$=0$};
  \node[above, font={\small\bfseries}]
    at (90:\R+0.52)  {ChartQA};
  \node[above right, font={\small\bfseries}, align=center]
    at (45:\R+0.35)  {Science\\QA};
  \node[right, font={\small\bfseries}]
    at (0:\R+0.52)   {VQAv2};
  \node[below right, font={\small\bfseries}, align=center]
    at (-45:\R+0.35) {MedXpert\\QA-MM};
  \node[below, font={\small\bfseries}]
    at (-90:\R+0.52) {PMC-VQA};
  \node[below left, font={\small\bfseries}, align=center]
    at (-135:\R+0.35){MMMU\\-Med};
  \node[left, font={\small\bfseries}]
    at (180:\R+0.52) {SLAKE};
  \node[above left, font={\small\bfseries}, align=center]
    at (135:\R+0.35) {VQA-RAD};
  \draw[c3b, line width=2pt, fill=c3b, fill opacity=0.18]
    (90:.960*\R)--(45:.330*\R)--(0:.440*\R)--(-45:.420*\R)
    --(-90:.300*\R)--(-135:.320*\R)--(180:.340*\R)--(135:.385*\R)--cycle;
  \draw[c7b, line width=2pt, fill=c7b, fill opacity=0.18]
    (90:.700*\R)--(45:.100*\R)--(0:.350*\R)--(-45:.320*\R)
    --(-90:.180*\R)--(-135:.205*\R)--(180:.340*\R)--(135:.315*\R)--cycle;
  \draw[crt, line width=2pt, fill=crt, fill opacity=0.18]
    (90:.800*\R)--(45:.110*\R)--(0:.520*\R)--(-45:.180*\R)
    --(-90:.200*\R)--(-135:.205*\R)--(180:.350*\R)--(135:.330*\R)--cycle;
  \draw[cri, line width=2pt, fill=cri, fill opacity=0.18]
    (90:.710*\R)--(45:.250*\R)--(0:.580*\R)--(-45:.240*\R)
    --(-90:.400*\R)--(-135:.230*\R)--(180:.360*\R)--(135:.295*\R)--cycle;
  \foreach \ang/\r in
    {90/.960, 45/.330, 0/.440, -45/.420, -90/.300, -135/.320, 180/.340, 135/.385}
    { \filldraw[fill=c3b, draw=white, line width=1pt] (\ang:\r*\R) circle (2.5pt); }
  \foreach \ang/\r in
    {90/.700, 45/.100, 0/.350, -45/.320, -90/.180, -135/.205, 180/.340, 135/.315}
    { \filldraw[fill=c7b, draw=white, line width=1pt] (\ang:\r*\R) circle (2.5pt); }
  \foreach \ang/\r in
    {90/.800, 45/.110, 0/.520, -45/.180, -90/.200, -135/.205, 180/.350, 135/.330}
    { \filldraw[fill=crt, draw=white, line width=1pt] (\ang:\r*\R) circle (2.5pt); }
  \foreach \ang/\r in
    {90/.710, 45/.250, 0/.580, -45/.240, -90/.400, -135/.230, 180/.360, 135/.295}
    { \filldraw[fill=cri, draw=white, line width=1pt] (\ang:\r*\R) circle (2.5pt); }
  \begin{scope}[shift={(0,-\R-2.0)}]
    \draw[gray!20, rounded corners=5pt, fill=gray!5]
      (-4.7,-0.52) rectangle (4.7,0.52);
    \draw[c3b, line width=2pt] (-4.25, 0.18)--(-3.45, 0.18);
    \filldraw[fill=c3b, draw=white, line width=1pt] (-3.85, 0.18) circle (2.5pt);
    \node[right, font=\scriptsize] at (-3.40, 0.18) {Qwen2.5-VL-3B};
    \draw[crt, line width=2pt] (-1.10, 0.18)--(-0.30, 0.18);
    \filldraw[fill=crt, draw=white, line width=1pt] (-0.70, 0.18) circle (2.5pt);
    \node[right, font=\scriptsize] at (-0.25, 0.18) {RL(text) ckpt};
    \draw[refcol, dashed, line width=1pt] (2.20, 0.18)--(2.90, 0.18);
    \node[right, font={\fontsize{6}{7}\selectfont}, refcol]
      at (2.95, 0.18) {BD\,$=$\,0};
    \draw[c7b, line width=2pt] (-4.25,-0.18)--(-3.45,-0.18);
    \filldraw[fill=c7b, draw=white, line width=1pt] (-3.85,-0.18) circle (2.5pt);
    \node[right, font=\scriptsize] at (-3.40,-0.18) {Qwen2.5-VL-7B};
    \draw[cri, line width=2pt] (-1.10,-0.18)--(-0.30,-0.18);
    \filldraw[fill=cri, draw=white, line width=1pt] (-0.70,-0.18) circle (2.5pt);
    \node[right, font=\scriptsize] at (-0.25,-0.18) {RL(image) ckpt};
  \end{scope}
\end{tikzpicture}%
}
\caption{
Blank Drop (BD $=$ $\mathrm{Acc}_{\mathrm{real}} - \mathrm{Acc}_{\mathrm{blank}}$)
across eight benchmarks for the four MedVLM families ($n{=}500$ per
benchmark). Each axis spans $[-0.07, +0.13]$. The dashed octagon
indicates the approximate $BD{=}0$ reference region; values inside
suggest language-shortcut behaviour, while values outside indicate that
removing the image reduces accuracy.
}
\label{fig:bd_radar}
\end{figure}
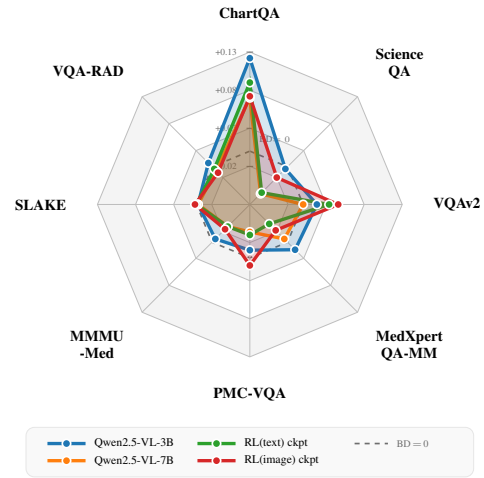

\paragraph{RL(image) text-prior signature.}
Even setting \ours{} aside, the RL(image) row of
Table~\ref{tab:phase2-full} is itself informative: this model achieves
its highest accuracy on \emph{no-image} prompts
(acc$_{\text{noimg}}{=}0.548$, the highest of any model) and shuffled
prompts ($0.488$), and its lowest accuracy on real images
(albeit marginally).  A model that performs almost as well without the
image as with it is not visually reliant.  This is the empirical
signature underlying the cross-domain shortcut pattern that
Figure~\ref{fig:bd_radar} visualises on PMC-VQA.

\subsubsection{\ours{} Per-Benchmark Behaviour}
\label{sec:results:coral}

Table~\ref{tab:sftlite-bench} reports per-benchmark item-level
bootstrap CIs.  The per-benchmark picture is consistent with
the macro story but more textured:

\begin{table}[t]
\centering
\small
\setlength{\tabcolsep}{3pt}
\caption{Per-benchmark \ours{} vs.\ Qwen2.5-VL-7B base, item-level bootstrap CIs.  \textbf{Bold} marks CIs
that exclude zero.}
\label{tab:sftlite-bench}
\begin{tabular}{@{}llrr@{}}
\toprule
Benchmark & Metric & $\Delta$ (95\% CI) & $P$ \\
\midrule
\multirow{3}{*}{PathVQA}
 & acc & $-0.050$ $[-0.170,\,+0.070]$ & 0.19 \\
 & \vrs{} & $+0.040$ $[-0.100,\,+0.180]$ & 0.70 \\
 & \vhr{} & \textbf{$-0.120$ $[-0.210,\,-0.030]$} & \textbf{0.99} \\
\midrule
\multirow{3}{*}{PMC-VQA}
 & acc & \textbf{$+0.150$ $[+0.040,\,+0.250]$} & \textbf{1.00} \\
 & \vrs{} & $+0.010$ $[-0.130,\,+0.160]$ & 0.55 \\
 & \vhr{} & $+0.010$ $[-0.070,\,+0.090]$ & 0.65 \\
\midrule
\multirow{3}{*}{SLAKE}
 & acc & $+0.110$ $[-0.020,\,+0.240]$ & 0.95 \\
 & \vrs{} & $+0.010$ $[-0.110,\,+0.130]$ & 0.55 \\
 & \vhr{} & $-0.070$ $[-0.150,\,+0.000]$ & 0.96 \\
\midrule
\multirow{3}{*}{VQA-RAD}
 & acc & $+0.060$ $[-0.060,\,+0.180]$ & 0.82 \\
 & \vrs{} & $+0.090$ $[-0.040,\,+0.220]$ & 0.90 \\
 & \vhr{} & \textbf{$-0.140$ $[-0.210,\,-0.080]$} & \textbf{1.00} \\
\bottomrule
\end{tabular}
\end{table}

\begin{itemize}
\item \textbf{PMC-VQA} shows the largest accuracy effect ($+15$\,pp,
  CI excludes zero), consistent with this being the training-data
  domain and with the richest HN pool ($\bar\mu = 0.974$).  \vhr{}
  is essentially flat, suggesting the accuracy gain comes from
  improved correct-answer extraction rather than from reduced
  shuffled-correct--real-wrong cases.
\item \textbf{SLAKE} shows a near-significant accuracy gain
  ($+11$\,pp, $P{=}0.95$) and a near-significant \vhr{} reduction
  ($-7$\,pp, $P{=}0.96$).
\item \textbf{VQA-RAD} shows the largest \vhr{} reduction in the entire
  evaluation ($-14$\,pp, CI excludes zero, $P{=}1.00$), even though
  its accuracy gain is modest ($+6$\,pp, n.s.). 
\item \textbf{PathVQA} shows a small accuracy regression
  ($-5$\,pp, n.s.) but the largest CI-excluding-zero \vhr{} reduction
  on a within-training-domain benchmark ($-12$\,pp).  This asymmetric
  pattern, loss of accuracy paired with reduced confident hallucinations, is what we would predict from a contrastive objective that
  systematically suppresses image-invariant predictions, including
  some that would have been correct.
\end{itemize}

\subsection{Cross-Domain Diagnostics ($n{=}500$)}
\label{sec:results:crossdomain}

The cross-domain pass uses $n{=}500$ items per benchmark across ten
benchmarks for the four-model MedVLThinker family.
We summarise and report the diagnostics that directly motivate
our framing (adversarial retrieval, occlusion, corruption,
dataset bias \& test-only solvability, synthetic calibration, and rationale stability).

\section{Analysis: Why \cgo{} Helps Accuracy and \vhr{}
but not \vrs{} or \bd{}}
\label{sec:analysis}

The contrast between the resolved (accuracy, \vhr{}) and unresolved
(\vrs{}, \bd{}) results is informative
about the mechanism of \cgo{}.  The contrastive term in
Eq.~\ref{eq:cgo} fires when the \emph{extracted answer} is the same on
the real image and on a CLIP-similar but label-different hard negative.
This term most directly targets the \vhr{} failure mode the model
arriving at the same wrong answer under image perturbation and we
observe a $-8.0$\,pp ($P{<}0.001$) macro reduction in \vhr{} that is
consistent across three of the four benchmarks (Table~\ref{tab:sftlite-bench}).
The accuracy gain follows partly from the same mechanism: by
penalising image-invariant errors at training time, \cgo{} shifts the
model away from the language-prior-only solution toward a solution
that uses the image at least to disambiguate visually confusable
neighbours.

\vrs{}, by contrast, requires the model to be doing more than
disambiguating against the hard negative: it must also be \emph{using
the image to reach the correct answer} when it would otherwise have
shuffled to a wrong answer.  Our contrastive term does not directly
optimise this counterfactual.  We therefore predict, and observe, a
positive but smaller and less resolved effect on \vrs{} at $n{=}400$.

The lack of an effect on \bd{} is consistent with the cross-domain
finding that medical benchmarks have very little headroom on the
blank/no-image contrast: if the question stem plus the system prompt
already supplies a strong prior, then replacing the image with a blank
does not change the answer distribution much, regardless of whether
the model is visually grounded.


\section{Conclusion}
\label{sec:conclusion}

Accuracy alone does not measure grounding. Across four medical VQA benchmarks, neither RL fine-tuned checkpoint achieves a statistically significant accuracy gain over Qwen2.5-VL-7B, and the RL(image) checkpoint peaks on \emph{no-image} prompts.  Hard-negative contrastive training (\cgo{})
on top of the same backbone gives a statistically resolved $+6.7$\,pp
accuracy gain (95\% CI $[+0.75,\,+12.5]$\,pp) and a $-8.0$\,pp \vhr{}
reduction ($P{<}0.001$), with a positive but unresolved \vrs{} trend at
$n{=}400$.  Two findings cut against an over-claim: $\imgsens$ on
extracted answers is \emph{worse} for \ours{} than for the base
(\S\ref{sec:results:phase2}), 
We treat \cgo{} therefore as a
demonstrably useful but incomplete corrective: it raises the cost of
language-prior shortcuts at training time without converting the model into a fully image-reliant reasoner.

\section*{Limitations}
\label{sec:limitations}

\paragraph{Evaluation scale.}
We evaluate $n{=}100$ fixed items per benchmark so that all models run
under the full six-condition protocol on identical samples. This yields
paired power for our primary claims the accuracy and \vhr{} effects
for \ours{} are statistically resolved while the secondary \vrs{} and
\bd{} trends remain directional at this $n$ and would benefit from
larger paired runs.

\paragraph{Metric and model scope.}
$\imgsens_{\text{pred}}$ is intentionally conservative: identical
extracted answers may reflect true invariance or distinct reasoning
paths converging on the same choice, and pairing it with rationale-level
attribution is left to future work. Our controlled comparison also
centres on Qwen2.5-VL and MedVLThinker-family 7B models; extending the
protocol to systems such as LLaVA-Med, BiomedGPT, and HuatuoGPT-Vision
would test generality but requires per-model prompt and answer-parsing
harmonisation.
\bibliography{custom}

\end{document}